# A Comprehensive Review on RNA Subcellular Localization Prediction

Cece Zhang[1+], Xuehuan Zhu[2+], Nick Peterson[3], Jieqiong Wang[4], Shibiao Wan[3,5*]

[1] *Department of Cell & Systems Biology, University of Toronto, ON, Canada.*
[2] *School of Engineering, University of California, Los Angeles, CA, United States.*
[3] *Department of Genetics, Cell Biology and Anatomy, University of Nebraska Medical Center, Omaha, NE, United States.*
[4] *Department of Neurological Sciences, University of Nebraska Medical Center, Omaha, NE, United States.*
[5] *Fred and Pamela Buffett Cancer Center, University of Nebraska Medical Center, Omaha, NE, United States.*

*Corresponding author. E-mail: swan@unmc.edu.
[+]These authors contributed equally to this work, and they should be regarded as co-first authors.

## Abstract

The subcellular localization of RNAs, including long non-coding RNAs (lncRNAs), messenger RNAs (mRNAs), microRNAs (miRNAs) and other smaller RNAs, plays a critical role in determining their biological functions. Understanding RNA localization sheds light on processes like gene expression regulation with spatial and temporal precision. However, traditional wet lab methods for determining RNA localization, such as in situ hybridization, are often time-consuming, resource-demanding, and costly. To overcome these challenges, computational methods leveraging artificial intelligence (AI) and machine learning (ML) have emerged as powerful alternatives, enabling large-scale prediction of RNA subcellular localization. This paper provides a comprehensive review of the latest advancements in AI-based approaches for RNA subcellular localization prediction, covering various RNA types and focusing on sequence-based, image-based, and hybrid methodologies that combine both data types. We highlight the potential of these methods to accelerate RNA research, uncover molecular pathways, and guide targeted disease treatments. Furthermore, we critically discuss the challenges in AI/ML approaches for RNA subcellular localization, such as data scarcity and lack of benchmarks, and opportunities to address them. This review aims to serve as a valuable resource for researchers seeking to develop innovative solutions in the field of RNA subcellular localization and beyond.

---

**Cece Zhang** is currently a BSc student in the Department of Cell & Systems Biology at the University of Toronto. Her current study focus is Bioinformatics.

**Xuehuan Zhu** received the M.ENG. degree in artificial intelligence from University of California, Los Angeles in Los Angeles, CA, United States, in 2023. He is seeking to be a phd student. HIs research interests include deep learning, artificial intelligence with healthcare, computer vision, and large language model.

**Nick Peterson** (Master's student, UNMC). Nick Peterson is currently a Master's student in the lab of Dr. Shibiao Wan at the University of Nebraska Medical Center. His current study focus is Bioinformatics. He graduated with a BSc in biology from the University of Nebraska Omaha (UNO) in May of 2024.

**Jieqiong Wang** received the BEng degree in telecommunication engineering from Wuhan University, China, in 2010 and the PhD degree in pattern recognition and intelligent systems from University of Chinese Academy of Sciences in 2015. She received the postdoctoral training in the Chinese University of Hong Kong, the Hong Kong Polytechnic University and the University of Pennsylvania. Her current research interests focus on machine learning and medical imaging processing, with the application in neuroscience and cancer. With over 10 years of experience in medical image processing/analysis, machine learning, and neuroscience, Dr. Wang has published numerous articles in high-impact journals such as Human Brain Mapping, European Radiology, and Brain Imaging and Behavior.

**Shibiao Wan** is currently an Assistant Professor and Co-Director for Bioinformatics and Systems Biology PhD Program at University of Nebraska Medical Center. With more than 15 years of experience in artificial intelligence, machine learning, bioinformatics, and computational biology, Dr. Wan has published 70+ research articles in journals like Advanced Science, Genome Research, Nature Communications, Science Advances, Circulation Research, and Nature Aging. Dr. Wan is the Editor-in-Chief for Current Proteomics and an editorial board member for Discover Oncology, BMC Bioinformatics, Briefings in Functional Genomics, PeerJ Computer Science, Heliyon, International Journal of Microbiology. He is also a reviewer for >70 journals including Nature Biotechnology, Nature Methods, and Nature Communications. Dr. Wan is a Senior Member of IEEE and ACM.

## 1. Introduction

RNA is a nucleic acid molecule that plays crucial roles in various cellular processes, including gene expression regulation, catalysis of biochemical reactions, and genetic information translation within cells. RNA is typically single-stranded and composed of a long chain of nucleotides. Each nucleotide contains a ribose sugar, a phosphate group, and one of four nitrogenous bases: adenine (A), guanine (G), cytosine (C), or uracil (U) (in contrast to thymine (T) in DNA). Multiple different categories of RNAs are found in the cell, including messenger RNA (mRNA), long non-coding RNA (lncRNA), microRNA (miRNA), etc. [1-2] RNA plays important roles in the regulation of both transcription and translation within different compartments of a cell [3-4]. In the nucleus, RNA is synthesized from DNA and serves functions such as splicing, capping, and polyadenylation in mRNA processing [5]. Mitochondrial-specific RNA encodes multiple proteins necessary for mitochondrial and metabolic functions [6]. RNA within the chloroplasts of plant cells promotes the expression photosynthesis related genes in the nucleus [7].

mRNA is a single-stranded RNA molecule that is complementary to a specific DNA strand within a genome [8]. mRNA serves as a crucial intermediary in the process of translating genetic information from our DNA into functional proteins [8]. mRNA is more mobile than DNA within cells and can exit the nucleus, allowing it to carry genetic information to the ribosomes in the cytoplasm and act as a template for protein synthesis during translation [9]. Structures within the mRNA also influence aspects of protein synthesis and gene expression regulation [10]. Before translation, mRNA strands undergo modifications to enhance stability (such as capping and polyadenylation) and to enable the selective expression of specific regions of the genetic code, allowing for the production of different proteins from the same gene through splicing [11]. mRNA is not evenly distributed throughout cells but is instead concentrated in specific cellular compartments [12,13]. A notable example is the asymmetric mRNA distribution in ascidian embryos, first revealed by Jeffery et al. [14], a significant finding that enhances our understanding of the mechanisms underlying cell differentiation in early embryonic development. The non-random organization associated with cytoskeletal proteins within the cytoplasm provides insight into a potential mechanism for measuring their concentration [15].

mRNA localization relies on three primary mechanisms: direct transport on the cytoskeleton through molecular motors, protection from degradation, and diffusion with local entrapment [16-18]. These mechanisms occur through the interaction of destination-specific proteins and adaptor proteins to form a ribonucleoprotein (RNP) complex. A pivotal determinant for the localization of mRNAs is the interaction between cis-acting signals, which are mainly located in the 3′ untranslated regions (UTRs) and 5′ ends of mRNA sequences [19-23] for facilitating the uneven distribution of specific transcripts. These regions, often referred to as "zip codes", are critical elements within the linear RNA sequence or structure, containing cis-regulatory elements that interact with trans-acting factors, predominantly RNA-binding proteins (RBPs). These RBPs further interact with other RBPs and RNA-binding domains (RBDs) [24,25]. These interactions are essential for the control of mRNA distribution within the cellular milieu.

lncRNAs are RNA molecules exceeding 200 nucleotides in length that, unlike mRNAs, do not encode proteins [26]. Instead, lncRNAs regulate various biological processes at transcriptional and post-transcriptional levels [27]. Their subcellular localization plays a crucial role in regulating gene expression through spatial and temporal control, which is essential for numerous cellular and developmental processes like regulated translation in highly polarized asymmetric cells, cell migration, maintenance of cellular polarity, orchestration of synaptic plasticity associated with long-term memory, assembly of protein complexes, asymmetric cell division, embryonic patterning, and cellular adaptation to stress [26,28-37]. LncRNAs can interact with chromatin-modifying complexes, altering chromatin structure and influencing gene expression by modifying transcriptional accessibility [26]. This interaction is vital in cellular differentiation and development, guiding cells toward specific fates during development and tissue formation [38]. In disease diagnosis and therapy, lncRNAs serve as important biomarkers for disease detection, prognosis, and treatment monitoring [39]. Dysregulated lncRNAs, such as MALAT1, H19, MEG3, and HOTAIR, have been linked to various cancers, while others like NBAT-1 are associated with poor cancer prognosis [40-43]. Furthermore, lncRNAs play pivotal roles in neurodegenerative diseases, such as amyotrophic lateral sclerosis (ALS), and inflammatory conditions, including inflammatory bowel diseases [44-48]. Their functional duality as oncogenes or tumor suppressors makes lncRNAs promising therapeutic targets [49].

Small RNAs, including miRNAs, are shorter than lncRNAs, typically under 200 nucleotides in length [50-52]. Unlike mRNAs, miRNAs are non-coding and regulate gene expression at the post-transcriptional level [53,54]. Aberrant expression of miRNAs has been implicated in diseases like cancer and COVID-19 [55,56], spurring research into miRNA-based diagnostic markers and therapeutic strategies using deep learning algorithms [57,58].

The subcellular localization of mRNA, lncRNA, and smaller RNAs such as miRNAs are integral to their functions. Proper localization of RNAs within cells is crucial for spatial and temporal regulation of gene expression, affecting cell differentiation, polarization, and development [26-28]. Disruptions in RNA localization are linked to diseases such as cancer [59-64], spinal muscular atrophy [65], neuronal dysfunction [59,66-71], and developmental disorders [72]. For instance, improper localization of lncRNAs, including MALAT1 and HOTAIR, has been associated with oncogenesis and metastasis [40-42]. These findings underscore the importance of understanding RNA localization to uncover mechanisms driving normal cellular processes and disease pathogenesis.

RNA subcellular localization prediction is inspired by parallel achievements in protein subcellular localization prediction. Although RNA and protein subcellular localization are closely related, their underlying factors differ. Protein localization is largely governed by targeting signals and protein trafficking pathways [86], whereas RNA localization depends more on cis-regulatory sequence elements, RNA structure, and interactions with RNA-binding proteins [89]. In the past decade, we have developed a series of AI/ML based methods for protein subcellular localization, including gene ontology-based like GOASVM [73], mGOASVM [74], R3P-Loc [75], mPLR-Loc [76] and HybridGO-Loc [77], interpretable features like FUEL-mLoc [78], mLASSO-Hum [79], SpaPredictor [80], and Gram-LocEN [81], and ensemble models like LNP-Chlo [82], EnTrans-Chlo [83] as well as membrane protein function prediction like Mem-mEN [84] and Mem-ADSVM [85]. Proteins and RNA interact in complex networks to regulate and perform cellular activities [86,87]. RNA serves as both a template and a regulatory molecule, whereas proteins often provide structural and catalytic roles that enable or modify RNA's functions [88,89]. Together, they establish the foundational processes of life, including gene expression, regulation, and cellular maintenance [90,91]. Advances in technology have led to rapid prediction of protein subcellular localization [92-94,196-197] and a comprehensive review can be found in [46]. Given the intricate relationship between RNA and proteins – such as protein synthesis relying on mRNA-encoded genetic information and RNA localization influencing disease mechanisms – many models have been developed to predict RNA subcellular localization, including for specific RNA types [95-97].

Many studies in the past decades have used wet lab methods to determine the subcellular localization of mRNA, lncRNA and miRNA. One common method is to use RNA fluorescent in situ hybridization (RNA-FISH) to detect subcellular localization based on which state-of-the-art technologies have been developed, including smFISH [98], MERFISH [99], seqFISH+ [100] and GeoMx DSP [101], to provide high resolution images of individual transcripts. However, these methods are time-consuming and labor-intensive. In recent years, advanced high throughput RNA sequencing methods such as APEX-RIP [102] CeFra-seq [103], and CLIP-seq [104] have been introduced to detect single RNA subcellular localization. These methodologies still have several flaws, including high complexity, inherent noise, and limitations in achieving high accuracy. Consequently, the limitations of both imaging-based and sequencing-based methods have driven the development of in-silico computational techniques, which offer a faster, more cost-effective, and accurate alternative for predicting the subcellular localization of mRNA, lncRNA, and miRNA [105-107].

Most in-silico computational methods are based on supervised machine learning approaches that require carefully designed features. These methods can be broadly categorized into three main types based on the features they utilize: (1) sequence-based methods, which use only amino acid sequences as input; (2) image-based methods, which rely on bio-image data to detect subcellular compartments; and (3) hybrid methods, which combine bio-image and sequence data to achieve higher accuracy by leveraging the strengths of each data type while mitigating their weaknesses. This paper reviews these approaches, discussing their methodologies, strengths, limitations, and applications in RNA subcellular localization prediction. Figure 1 provides an overview of different machine learning approaches and the main steps involved. The essential components of machine learning approaches to RNA subcellular localization are inputs (nucleotide sequences in FASTA format, bio-images), feature extraction (including sequence-based, image-based, and hybrid approaches), prediction model using traditional machine learning and deep learning approaches for sequence-based and image-based models and fusion algorithms for the hybrid model, and finally RNA subcellular localization to assign a particular RNA to a single localization or multiple localizations.

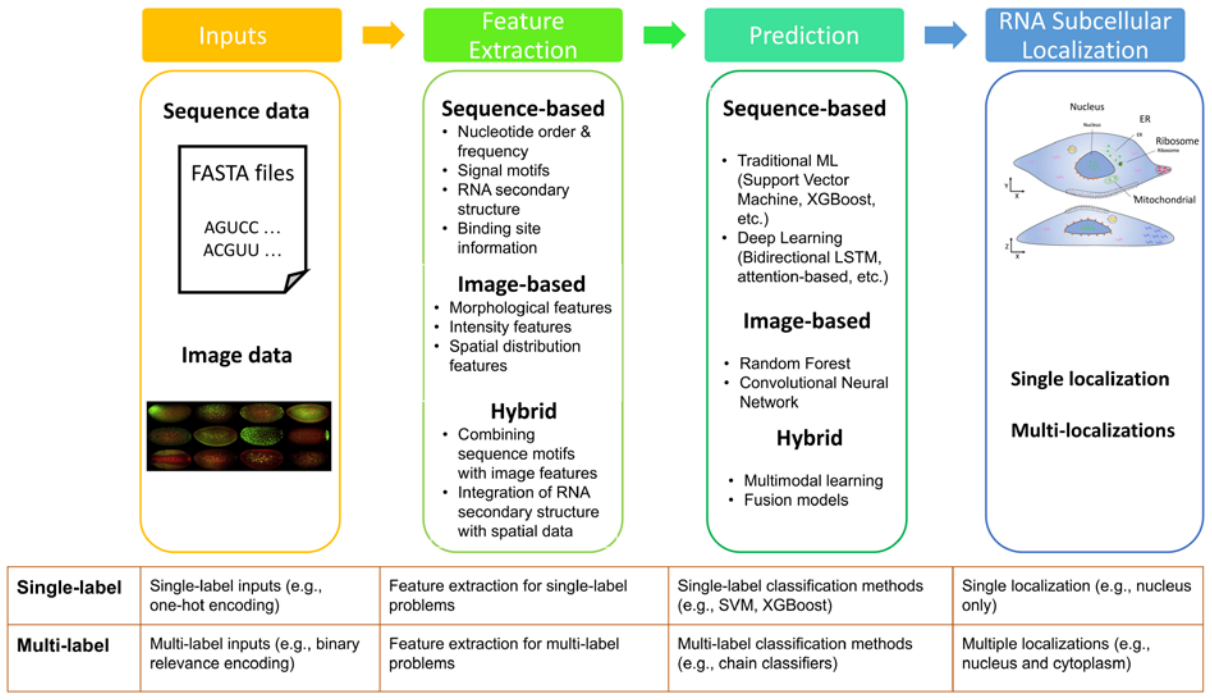

**Figure 1 The overview of machine learning approaches to RNA subcellular localization.**

The inputs are usually nucleotide sequences in FASTA format, bio-images (illustrative image from [13]), or a hybrid of both. The inputs need to be prepared to accommodate the single-label or multi-label task formulation. Typical features used in sequence-based approaches include sequence composition, nucleotide frequency distribution, signal motifs, etc. Image-based approaches use morphological features, spatial distribution features, etc. Hybrid approaches use a combination of both sequence and image features. Typical prediction algorithms include traditional machine learning and deep learning approaches for sequence-based and image-based models, fusion algorithms for the hybrid model, etc. The choice of prediction algorithms is also tailored to the single-label or multi-label task formulation. For example, a chain of classifiers is often applied to multi-label localization. Prediction outputs for an RNA can be one localization in the single-label setting or multiple localizations in the multi-label setting (illustrative image from [173]).

In the rest of the paper, Section 2 introduces the common features and algorithms in sequence-based methods. Section 3 covers image-based methods as well as hybrid methods that leverage both image and sequence-based data. In Section 4, existing challenges and future directions are explored and our expectations outlined. Section 5 concludes the paper.

## 2. Sequence-based Methods

In this section, we introduce typical RNA subcellular localization methods based on sequence data. Figure 2

summarizes the general features and algorithms used in sequence-based methods. First, training data are prepared by sequence refining with techniques such as Cluster Database at High Identity with Tolerance (CD-HIT) [108] to improve quality. Data imbalance is addressed with techniques like Synthetic Minority Over-sampling Technique (SMOTE) [109]. Then sequence features are extracted with techniques like K-mer [110-112], Pseudo K-tuple Nucleotide Composition (PseKNC) [113-115], Z-curve [116], etc. before feature selection is applied. Optimal and relevant features are selected with techniques like binomial distribution [117], Analysis of Variance (ANOVA) [28], Incremental Feature Selection (IFS) [117], etc. Finally, the features are used to train machine learning models with different algorithms, including conventional machine learning algorithms, deep learning algorithms, or ensemble methods.

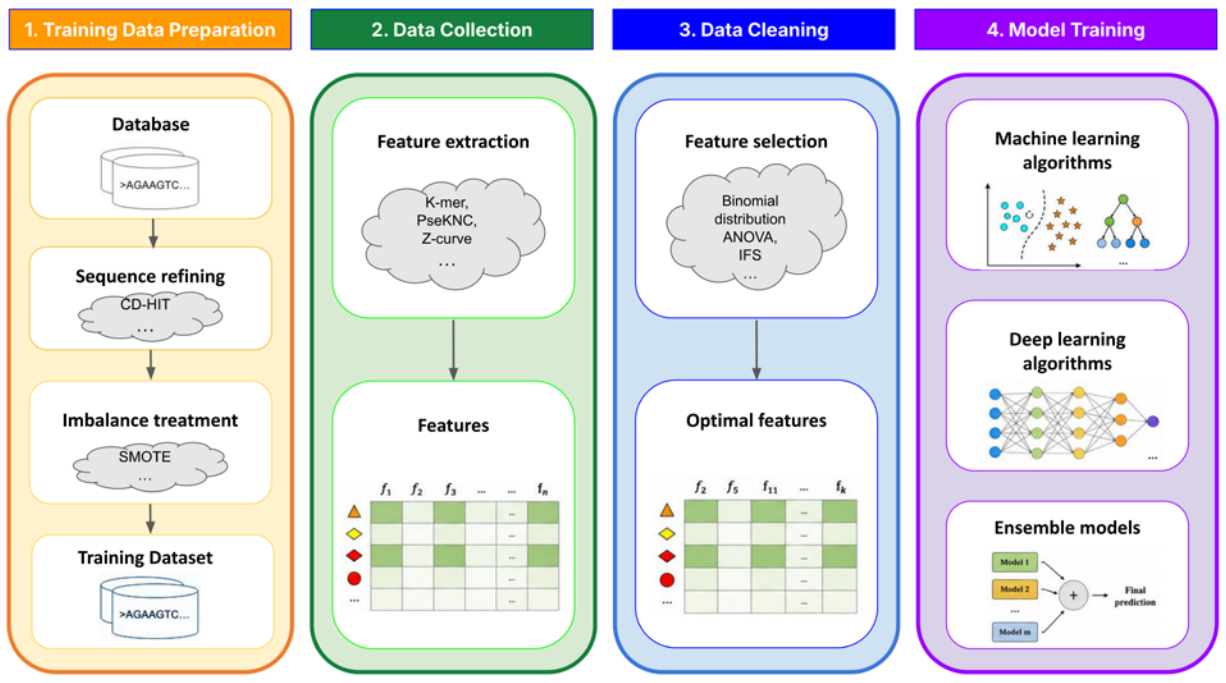

**Figure 2 Sequence-based approaches for predicting RNA subcellular location.**
Training data preparation involves techniques aimed at enhancing sequence quality for further use. Data collection focuses on extracting the desired features required for different models. Data cleaning is then performed to select optimal and relevant features from these desired features. The cleaned data is then used for model training. CD-HIT: Cluster Database at High Identity with Tolerance. SMOTE: Synthetic Minority Over-sampling Technique. PseKNC: Pseudo K-tuple Nucleotide Composition. ANOVA: Analysis of Variance. IFS: Incremental Feature Selection.

### *2.1. Sequence-based Features*

Many state-of-the-art approaches on RNA subcellular localization are sequence-based because the RNA sequences are easier and cheaper to obtain compared to other features. Sequence features can exhibit various characteristics like physicochemical properties [118], nucleic acid composition [119,120], and 3D representation [121,122]. Numerous feature extraction methods are used to analyze RNA nucleotides, secondary structures, RNA-binding motifs, and genomic loci. Additionally, some models employ self-defined features generated through in silico methodologies, such as large language models [123], to make predictions. Typically, feature selection techniques are used to identify which features contribute most to different subcellular compartments. These models [28,117,124] aim to predict the subcellular locations of mRNAs or lncRNAs by analyzing the correlations between these locations and the information encoded in nucleotide sequences.

**Table 1** A summary of sequence-based features for RNA subcellular localization.

| Feature | Description | References |
|---|---|---|
| K-mer | A contiguous subsequence of length K from a longer sequence of nucleotides | [110-112] |
| PseKNC | K-mer + distance or correlation values between K-mers at different positions in the sequence | [113-115] |
| PseEIIP | Integrating the Electron-Ion Interaction Potential (EIIP) values of nucleotides with the conventional k-tuple nucleotide composition to create a more informative representation of RNA sequences | [125-129] |
| One-hot encoding | Representing categorical data, such as nucleotide or amino acid sequences, in a vectorial format of numerical values with only a single 1 and 0's in every other place | [130-132] |

*Note*: PseKNC: Pseudo K-tuple Nucleotide Composition; PseEIIP: Pseudo Electron-Ion Interaction Potential.

Specifically, Table 1 summarizes different kinds of sequence-based features used for RNA subcellular localization. One of the most commonly used features is composition-based features. K-mer [110] is a typical method employed in this context, where a long sequence of nucleotides is analyzed as contiguous subsequences of length K. K-mers facilitate faster sequence comparisons by breaking down sequences into smaller, more manageable pieces, which is particularly useful in large-scale genomic data analysis. Building on this concept, Yuan et al. [111] proposed a model utilizing k-mers of lengths 3, 4, and 5 to identify sequence features and their associated RBPs that contribute to lncRNA subcellular localization. Yan et al. [112] developed RNATracker, one of the first computational predictors of mRNA subcellular localization using k-mers, to identify candidate cis-regulatory regions in strongly localized transcripts. However, k-mers may lack contextual information beyond their immediate sequence. To complement this, sequence order information is another type of features which are widely used.

Sequence order information is crucial in bioinformatics and genomics as it captures the arrangement and dependencies between nucleotides in a sequence, which cannot be achieved through k-mer analysis alone. PseKNC [113] is used to capture this information by introducing distance or correlation between k-mers at different positions within the sequence. Garg [114] applied PseKNC with k-values of 3, 4, and 5 to predict five subcellular localizations of eukaryotic mRNAs using cDNA/mRNA sequences in a novel model. Su [115] additionally proposed a model incorporating PseKNC to predict the subcellular localization of lncRNAs. PseKNC has certain drawbacks however, including overfitting, limited generalization without feature selection methods, and reduced interpretability. In miRNAs subcellular localization prediction, a novel approach called kmerPR2vec, which is a fusion of positional information of k-mer and k-mer embedding, has been proposed to carry more semantic information and differentiation ability [133].

In addition, physicochemical pattern features are descriptors used in bioinformatics and computational biology to capture the inherent physical and chemical properties of biological sequences like DNA, RNA, and proteins. These features help to understand the functional and structural aspects of sequences by considering properties like hydrophobicity, charge, mass, polarity, and others. Pseudo Electron-Ion Interaction Potential (PseEIIP) is a computational method to represent nucleotide sequences. It integrates the EIIP values of nucleotides with the conventional k-tuple nucleotide composition, creating a more informative representation of RNA sequences by incorporating physicochemical information into sequence analysis. Many models [125-129] utilize physicochemical pattern features to identify potential drug targets and predict the subcellular localization of mRNA and lncRNA by incorporating these features into the input data.

To improve the prediction performance, some models combine different types of features in term of prediction. For example, MMLmiRLocNet [134] gathered multi-perspective sequence representations by combining lexical features based on k-mer physicochemical properties, syntactic features derived from word2vec embeddings, and semantic representations created using pre-trained feature embeddings.

As a popular data representation technique in natural language processing, one-hot encoding is widely adopted in bioinformatics for representing categorical data like nucleotide or amino acid sequences in a vectorial format of numerical values. For example, DNA and RNA sequences are composed of four types of nucleotides (A, T/U, C, G). One-hot encoding represents each nucleotide as a binary vector of length four, namely (1, 0, 0, 0) for A, (0, 1, 0, 0) for T/U, (0, 0, 1, 0) for C, and (0, 0, 0, 1) for G. In such a vector there is a single 1 and 0's in every other place, hence the name "one-hot". This feature representation method was adopted by [41], where the RNA primary structure is represented by 4 bits as shown above, the RNA secondary structure is represented by 6 bits, and a joint primary-secondary representation used 4 x 6 = 24 bits. Compared with the other features, one-hot encoding is very simple and widely applicable. The main disadvantages for one-hot encoding are high dimensionality from long sequences and sparsity due to the many zeroes which may result in increased computational load and memory usage. One-hot encoding is typically used as input for deep learning algorithms [130-132] where no biologically inspired feature engineering is needed.

After feature extraction, feature selection is widely applied in various models as features often contribute differently across different compartments. Feature selection plays a vital role in enhancing model performance, reducing complexity and noise while improving model interpretability. One approach involves using a predictive model to evaluate combinations of features and select the best-performing subset. For example, Tang et al. [125] employed the sequential forward search (SFS) strategy to identify optimal feature subsets after removing highly correlated features and ranking the remaining ones. Zhang et al. [117] introduced an IFS strategy with a binomial distribution score. To address the issue of redundancy, the minimal-redundancy-maximal-relevance (mRMR) criterion was used in the model. Another approach for feature selection is based on the statistical properties of the data. Zhang et al. [28] used a combination of the binomial distribution and ANOVA to filter out irrelevant features, while Ahmad et al. [124] employed the Pearson correlation coefficient to identify the most correlated features.

Before feature extraction, Zhang and Qiao [135] employed non-negative matrix factorization (NMF) to analyze images, followed by Kullback-Leibler divergence-based non-negative matrix factorization (KLNMF) to extract features with a high contribution rate. Similar statistics-based data filtering methods are adopted by Fan et al. [128], who utilized Variance Threshold to filter out features whose variances did not meet the specified threshold.

### *2.2. Sequence-based Algorithms*

After feature extraction and selection, classification algorithms are important to make final predictions of RNA subcellular localization. As the number of features increases, model complexity, sensitivity to parameters, and computational demand also rise, requiring different levels of AI/ML algorithms ranging from conventional machine learning approaches to complex deep learning models. In addition to discussing the development of these models, we will explore techniques for handling multi-label mRNA, lncRNA, and miRNA, as well as methodologies for addressing imbalanced data problems.

**Table 2** A summary of sequence-based methods for RNA subcellular localization.

| **Method** | **Algorithm** | **Localizations** | **Dataset Size** | **Metrics** | **Single / Multiple Labels** | **Reference** | **Year** |
|---|---|---|---|---|---|---|---|
| MulStack | RF, CNN, BiLSTM | Exosome, membrane, cytosol, ribosome, endoplasmic reticulum, and nucleus for mRNA | 17,023 | AUC | Multiple | [136] | 2024 |
| UMSLP | CatBoost, XGBoost, DT, GNB, MLP | Nucleus, cytoplasm, extracellular region, mitochondria, endoplasmic reticulum for mRNA | 14,909 | Precision, Recall, Accuracy, F1 | Single | [116] | 2024 |
| DeepLocRNA | Deep learning with attention | Nucleus, exosome, cytosol, cytoplasm, ribosome, membrane, endoplasmic reticulum, microvesicle, and mitochondrion for mRNA | 60,850 | Precision, Recall, F1, AUROC, AUPRC | Multiple | [130] | 2024 |
| Zuckerman & Ulitsky | RF | Cell, cytosol, nucleus for LncRNA | 8,967 | Precision, Recall, AUC | Single | [137] | 2024 |
| Zhang et al. | BERT | Nucleus, exosome, cytoplasm, ribosome, cytosol, extracellular vesicle for LncRNA | 907 | Precision, Recall, F1, AUROC, AUPRC | Multiple | [138] | 2024 |
| PreSubLncR | CNN, BiLSTM | Cytoplasm, exosome, nucleus, and ribosome for LncRNA | 7,69 | Precision, Recall, Accuracy, F1 | Single | [139] | 2024 |
| GATLncLoc+ C&S | GNN | Cytoplasm, cytosol, exosome, nucleus, and ribosome for LncRNA | 1,212 | Precision, Recall, Accuracy, F1, Sensitivity, Specificity, MCC | Single | [140] | 2024 |
| Deng | GNN | Cytoplasm, cytosol, exosome, nucleus, and ribosome for LncRNA | 1,212 | Precision, Recall, Accuracy, F1 | Single | [141] | 2024 |
| MMLmiRLocNet | Neural Network | Cytoplasm, exosome, nucleus, microvesicle for miRNA | 538 | Accuracy, F1, AUC | Multiple | [134] | 2024 |
| PMiSLocMF | Neural Network with Self-attention | Cytoplasm, exosome, nucleolus, nucleus, extracellular vesicle, microvesicle, mitochondrion for miRNA | 1,041 | AUROC, AUPRC, Accuracy | Multiple | [174] | 2024 |
| MGFmiRNAloc | GCN | Cytoplasm, exosome, nucleus, microvesicle, circulating, mitochondrion for miRNA | 1,048 | Accuracy, Precision, Recall, F1 | Single | [176] | 2024 |
| DAmiRLocGNet | GCN, Autoencoder | Cytoplasm, exosome, nucleolus, nucleus, extracellular vesicle, microvesicle, mitochondrion for miRNA | 1,041 | Precision, Recall, Specificity, FPR, AUROC, AUPRC | Single | [177] | 2024 |
| RNAlight | LightGBM | Nucleus and cytoplasm for mRNA | 8,972 | Accuracy, Sensitivity, Specificity, MCC, F1, AUROC | Single | [111] | 2023 |
| MSLP | Ensemble (RF, XGBoost, CatBoost, DT, SVM, GNB) | Cytoplasm, endoplasmic reticulum, extracellular, mitochondria and nucleus for mRNA | 22,875 | Precision, Accuracy, F1, Sensitivity, Specificity, MCC | Single | [127] | 2023 |
| MRSLpred | CNN, XGBoost | Ribosome, cytosol, endoplasmic reticulum, membrane, nucleus, and exosome for mRNA | 41,835 | Accuracy, F1, Sensitivity, Specificity, MCC | Multiple | [132] | 2023 |
| Allocator | MLP, Graph-based | Nucleus, exosome, cytosol, ribosome, membrane, and endoplasmic reticulum for mRNA | 17,928 | Accuracy, Precision, Coverage, One-error, Ranking loss, Hamming loss | Multiple | [142] | 2023 |
| NN-RNALoc | Neural Network | Cytosol, Insoluble, Membrane, Nuclear for mRNA | 11,180 | Precision, Recall, F1, Accuracy, MCC | Single | [143] | 2023 |
| GraphLncLoc | GCN | nucleus, cytoplasm, ribosome, exosome for LncRNA | 769 | Precision, Recall, F1 | Single | [144] | 2023 |
| GM-lncLoc | GCN | Nucleus, exosome, cytoplasm, ribosome, cytosol for LncRNA | 3,523 | Precision, Recall, Accuracy, F1 | Single | [145] | 2023 |
| lncLocator-imb | CNN, RNN | Cytoplasm, nucleus for LncRNA | 167,136 | F1, AUC | Single | [129] | 2023 |
| LightGBM-LncLoc | LightGBM | Nucleus, exosome, cytoplasm, ribosome, cytosol for LncRNA | 924 | Accuracy, F1, AUC | Single | [146] | 2023 |
| LncLocFormer | Transformer | Nucleus, cytoplasm, chromatin, and insoluble cytoplasm for LncRNA | 811 | Precision, Recall, F1, AUC | Multiple | [147] | 2023 |
| DlncRNALoc | SVM | Cytoplasm, cytosol, exosome, nucleus, and ribosome for LncRNA | 2,036 | Accuracy, Recall, F1, Sensitivity, Specificity, MCC | Single | [148] | 2023 |
| SGCL-LncLoc | GCN | Cytoplasm, nucleus for LncRNA | 4,485 | Precision, Recall, F1, Accuracy, MCC | Single | [149] | 2023 |
| LncDLSM | CNN | Cytoplasm, cytosol, exosome, nucleus, ribosome, etc. (from the NONCODE database) for LncRNA | 170,000 | Accuracy, F1, AUC, Sensitivity, Specificity | Single | [150] | 2023 |
| EL-RMLocNet | LSTM | Pseudopodium, nucleolus, nucleus, cytosol, mitochondrion, ribosome, endoplasmic reticulum, exosome, microvesicle, and cytoplasm for mRNA | 38,423 | Accuracy, Precision, Coverage, Ranking loss, One error | Multiple | [151] | 2022 |
| Clarion | XGBoost | Chromatin, cytoplasm, cytosol, exosome, membrane, nucleolus, nucleoplasm, nucleus and ribosome for mRNA | 36,971 | Accuracy, Precision, Coverage, One-error, Ranking loss, Hamming loss | Multiple | [152] | 2022 |
| DeepLncLoc | Text CNN | Nucleus, exosome, cytoplasm, ribosome, cytosol for LncRNA | 857 | Precision, Recall, F1, Accuracy | Single | [153] | 2022 |
| TACOS | Tree-based (RF, ERT, XGBoost) | Cytoplasm, nucleus for LncRNA | 127,294 | Accuracy, MCC, Sensitivity, Specificity | Single | [154] | 2022 |
| MiRLoc | Random Walk with Restart | Cytoplasm, exosome, nucleolus, nucleus, extracellular vesicle, microvesicle, mitochondrion for miRNA | 3,877 | Precision, Recall, FPR | Single | [106] | 2022 |
| iLoc-miRNA | BiLSTM | Exosome, circulating, extracellular vesicle, microvesicle, axon, cytoplasm, nucleus, mitochondrion for miRNA | 1,665 | Precision, Recall, F1, MCC, AUROC, AUPRC, HL, Jaccard | Multiple | [173] | 2022 |
| mRNALocater | Ensemble (LightGBM, XGBoost, CatBoost) | Cytoplasm, endoplasmic reticulum, extracellular, mitochondria and nucleus for mRNA | 14,909 | Precision, Recall, F1, Accuracy | Single | [125] | 2021 |
| SubLocEP | LightGBM | Cytoplasm, endoplasmic reticulum, extracellular, mitochondria and nucleus for mRNA | 31,790 | Precision, Recall, F1, Accuracy | Single | [126] | 2021 |
| DM3Loc | CNN, BiLSTM | Exosome, membrane, cytosol, ribosome, endoplasmic reticulum, and nucleus for mRNA | 17,023 | AUROC, AUPRC, MCC | Multiple | [131] | 2021 |
| mLoc-mRNA | RF, EN | cytoplasm, endoplasmic reticulum, cytosol, exosome, mitochondrion, nucleus, pseudopodium, posterior, ribosome for mRNA | 11,827 | Accuracy, F1, MCC, Sensitivity, Specificity, AUROC, AUPRC | Multiple | [155] | 2021 |
| Yi & Adjeroh | CNN | Nucleus, exosome, cytoplasm, ribosome, cytosol for LncRNA | 2,534 | MCC, Sensitivity, Specificity | Single | [156] | 2021 |
| Locate-R | SVM | nucleus, cytoplasm, ribosome, exosome for LncRNA | 3,011 | Accuracy, MCC, Sensitivity, Specificity | Single | [124] | 2020 |
| LncLocation | SVM, RF, LR, XGBoost, LightGBM | nucleus, cytoplasm, ribosome, exosome for LncRNA | 655 | Precision, Recall, Accuracy, F1 | Single | [128] | 2020 |
| lncLocPred | LR | Nucleus, cytoplasm, ribosome, and exosome for LncRNA | 811 | Precision, Recall, Accuracy, F1 | Single | [157] | 2020 |
| miRNALoc | SVM | Axon, circulating, cytoplasm, exosome, extracellular vesicle, microvesicle, mitochondrion, nucleus for miRNA | 691 | AUROC, AUPRC, Sensitivity, Specificity, F1, MCC | Single | [107] | 2020 |
| RNATracker | CNN, LSTM | Cytosol, nuclear, membranes, insoluble, endoplasmic reticulum, mitochondria for mRNA | 11,373 | AUROC, AUPRC | Single | [112] | 2019 |
| DeepLncRNA | DNN | Nuclear, cytosolic for LncRNA | 93 | F1, Accuracy, Sensitivity, Specificity, AUC, MCC | Single | [158] | 2018 |
| lncLocator | SVM, RF | Nucleus, exosome, cytoplasm, ribosome, cytosol for LncRNA | 612 | Recall, Accuracy, F1 | Single | [159] | 2018 |
| DeepLNC | DNN | Cytoplasm, cytosol, exosome, nucleus, ribosome, etc. (from the LNCipedia and RefSeq databases) for LncRNA | 60,000 | Precision, Accuracy, Sensitivity, Specificity | Single | [160] | 2016 |

*Note*: RF: Random Forest; CNN: Convolutional Neural Network; BiLSTM: Bi-directional Long short-term memory; CatBoost: Categorical Boost; XGBoost: eXtreme Gradient Boosting; DT: Decision Tree; GNB: Gaussian Naive Bayes; MLP: Multilayer Perceptron; BERT: Bidirectional Encoder Representations from Transformers; GNN: Graph Neural Network; LightGBM: Light Gradient Boosting Machine; GCN: Graph Convolutional Network; RNN: Recurrent Neural Network; SVM: Support Vector

Machine; LSTM: Long short-term memory; ERT: Extremely Randomized Trees; EN: Elastic Net; LR: Logistic Regression; DNN: Deep Neural Network.

Table 2 shows a comprehensive list of sequence-based methods ordered chronologically according to their publication years. In the following, we give more details to the algorithms adopted by those methods.

*2.2.1 Conventional Machine Learning Approaches*

In several related research areas, such as protein subcellular localization, conventional machine learning classification is very popular due to its computational efficiency and interpretability [128,161-164]. Similarly, in RNA subcellular localization, traditional machine learning methods such as Support Vector Machines (SVM), Logistic Regression [157], tree-based methods [154], and Random Forests (RF) are widely utilized. These methods are favored for their substantial computational cost savings, allowing for fast predictions when handling low-level features.

For example, Fu et al. [148] proposed a discrete wavelet transform (DWT) feature extraction model to process the physicochemical property matrix of 2-tuple bases before applying SVM and optimizing the feature information using the local Fisher discriminant analysis (LFDA) algorithm. Zuckerman et al. [137] employed a Random Forest classifier to train and test different cell lines in humans and mice. While these models are robust, they often show inconsistent performance across different RNA localization settings and limited effectiveness in complex scenarios.

*2.2.2 Neural Network-based Deep Learning Approaches*

*2.2.2.1 Grid-based and Sequential Architectures*

In recent years, deep learning methods, particularly those involving neural networks, have gained popularity in lncRNA/mRNA subcellular localization prediction research due to their superior performance compared to conventional machine learning methods. Convolutional Neural Networks (CNNs) are widely used in this area of research, processing features embedded in sequences. Zeng et al. [153] introduced an innovative approach using TextCNN, a powerful deep learning network structure commonly used for text classification. They proposed a new subsequence embedding method where lncRNA sequences are divided into non-overlapping consecutive subsequences. Patterns from each subsequence are then extracted and combined to create a comprehensive representation that preserves sequence order information. To input data into TextCNN, the word2vec word embedding technique, widely used in natural language processing (NLP), was employed.

To handle the long-range dependencies inherent in RNA sequences, Long Short-Term Memory (LSTM) [171] networks are another widely used technique in deep learning research. Bidirectional LSTM layers and attention mechanisms are often combined with CNNs to achieve higher accuracy in prediction tasks [112,139,172]. This has been used in distinguishing extracellular miRNAs from intracellular miRNAs [173]. Wang et al. [174] used self-attention and fully connected layer to predict miRNAs with different concatenated features including miRNA sequence features converted from sequence similarity network by node2vec. In another study, Wang et al. [131] proposed a multiscale CNN with a multi-head self-attention architecture and analyzed sequence binding motifs extracted from CNN filters. In addition, a CNN with multi-head attention was used by Wang et al. [130] to extract RBP binding signals and utilize Integrated Gradients (IG) scores to analyze motifs. Liu et al. [129] combined CNNs with Gated Recurrent Units (GRU) to address the vanishing gradient and long-term memory problems. During training, they used a label-distribution aware margin (LDAM) loss function to tackle the class imbalance problem. Finally, they introduced the SHAP framework [175] to visualize the model using a physical and chemical property matrix via Normalized Moreau-Broto auto-cross correlation (NMBACC) and sequence order information via word2vec. Word embedding technologies such as word2vec are incorporated in many deep learning training pipelines, for instance, Zhang et al. [138] proposed using a pretrained BERT model after the word embedding step. Furthermore, Zeng et al. [147] used transformer blocks with localization-specific attention. Deep Neural Networks (DNNs) are a common choice for handling high-level features [158,160]. Babaiha et al. [143] proposed an Artificial Neural Network (ANN) to assign probabilities of mRNA belonging to specific compartments based on their distance-based representations. This approach addressed the issue that when feature vectors become extremely large and sparse, models may become memory inefficient and suffer from low performance.

*2.2.2.2 Topology-based and Graph Architectures*

A growing trend involves reformulating RNA sequences as graph problems to capture structural or semantic relationships. Some models transform lncRNA sequences into de Bruijn graphs [144]. For instance, Li et al. [149] applied graph convolutional networks (GCNs) to extract high-level features from these graphs after the transformation. Similarly, Deng et al. [140,141] utilized GCNs to predict lncRNA by constructing edges based on cosine similarity, measuring the similarity of features between nodes. This approach was also used with channel attention and spatial attention mechanisms (CBAM) in miRNAs subcellular localization prediction [176]. Furthermore, Deng et al. [177] combined GCN with autoencoder to gather information in neighboring nodes and implicit information of structure based on miRNA sequence semantics information extracted from sequence similarity networks, miRNA–disease association information, and disease semantic information. To address the issues of data imbalance and overfitting, a weighted graph attention network (R-GAT) combined with graph attention mechanisms and weighted loss was proposed. In comparison Cai et al. [145] integrated GCN with Model-Agnostic Meta-Learning (MAML) and a protein sequence similarity network (SSN) to achieve high prediction accuracy. Malik et al. [151] meanwhile employed the GeneticSeq2Vec approach, a k-hop neighborhood relation-based statistical representation scheme for RNA sequences, to generate graphs, which were then fed into an LSTM layer. Li et al. [142] discovered that a Multilayer Perceptron (MLP) with an attention layer and the Adam optimizer achieved the highest accuracy when applied to k-mer and K-spacer nucleic acid pair composition (CKSNAP) features.

Deep learning methods typically demonstrate exceptional performance with high-dimensional feature input. When combined with traditional machine learning approaches, those models exhibit enhanced robustness and flexibility in RNA subcellular localization prediction, achieving higher accuracy and improved generalization. These approaches may require a larger dataset for training than traditional machine learning approaches, however.

#### *2.2.3 Addressing Multi-label and Imbalanced Data*

Although it is tempting to consider RNA subcellular localization prediction as a single-label classification problem - i.e., predicting one RNA to be in one particular subcellular compartment - the problem is inherently multi-label as on most occasions, RNAs may co-localize or move between two or more subcellular compartments. The most common localizations are exosome, membrane, cytosol, ribosome, endoplasmic reticulum (ER), and nucleus [131,132,136,142] though other localizations such as mitochondrion [151], ribosome [155], and microvesicle [130] were also reported. As multi-label classification is more challenging than single-label classification, the above-mentioned traditional machine learning algorithms are usually not the best choice when used alone. For instance, MulStack [136] is an ensemble learning model that adopts both random forest and deep learning algorithms. A similar strategy is adopted by Liu et al. in a 2014 study [113] as a hybrid approach that uses both XGBoost and convolutional neural networks. In many cases [136,142,147,151], deep neural networks with the attention mechanism were usually preferred because they can adaptively weigh the importance of specific sequence regions and capture complex, non-linear interactions that underlie spatial distribution. Also, inspired by success of data transformation in natural language processing, L2S-MirLoc [165] converted multi-label miRNA subcellular localization problem into multi-class problem with different machine learning approaches.

Using traditional models also has drawbacks, such as inconsistent performance across different compartments and poor effectiveness in complex scenarios, especially in multi-location predictions. In miRNA subcellular localization prediction, multi-label location prediction is a common scenario [166,167]. To address these issues, many ensemble methods are employed. Initially, LightGBM, as an ensemble model, was often used independently for prediction tasks [111,146]. This approach is similar to XGBoost [152], both of which are gradient boosting frameworks for supervised learning tasks. Subsequently, some models integrated these methods with others, such as CatBoost, SVM, and RF for a single prediction task [125,128] or combined them with sequence-based and physicochemical-based models [126]. Additionally, some models utilized the one-versus-all (OVA) approach [127] to predict multiple positions. To leverage the advantages of both traditional and deep learning approaches, some models such as CNN, MLP, and RF [124,127,132,136] incorporated a combination of methodologies.

Traditional machine learning methods generally perform well on well-designed features. However, when data are unevenly distributed across different subcellular localizations, the prediction performance will vary significantly. To address this challenge and enhance model stability, techniques like SMOTE [109] are often employed. Other commonly used methods include Random Over-sampling (ROS) [168], Supervised Over-Sampling (SOS) [159], binomial distribution-based filtering [128], and recursive feature elimination (RFE) based on Autoencoder [128]. GP-HTNLoc [169] chose separate training approaches for head and tail location labels to solve the problem of data imbalance and scarcity. Also, there is a new dataset established based on previous work [170]. To mitigate the data imbalance and overfitting problems arising from sequence-to-graph transformations, a weighted graph attention network (R-GAT) incorporating graph attention mechanisms and a weighted loss was proposed.

### *2.3 Sequence-based Methods Progression*

The landscape of computational models for predicting RNA subcellular localization has evolved dramatically over recent years. Chronologically, there is a clear paradigm shift from traditional, straightforward machine learning algorithms – such as SVM [74] and RF [155] – toward increasingly sophisticated deep learning architectures and ensemble methods. Simultaneously, the field has transitioned from treating localization as a simplistic single-label problem to tackling it as a more realistic multi-label classification task. This shift acknowledges the biological reality that RNA molecules are dynamic and often reside in multiple cellular localizations simultaneously rather than being restricted to a single localization.

This methodological evolution is mirrored by advancements in feature engineering strategies. Early approaches relied heavily on hand-crafted features, including *k*-mer frequencies, PseKNC, and various physicochemical properties [113, 114]. While pioneering, these classical representations, particularly *k*-mer frequency and one-hot encoding, suffered from several intrinsic limitations. They largely ignore the sequential order of nucleotides, meaning crucial positional information is lost. Furthermore, performance is highly sensitive to the choice of k, creating a dilemma where a small *k* fails to capture sufficient sequence information, whereas an excessively large *k* exponentially inflates the feature space, leading to computational waste and severe model overfitting [153]. Additionally, one-hot encoding treats nucleotides as independent entities, overlooking the biochemical and evolutionary relationships between different bases [112].

To overcome these bottlenecks, modern architectures leverage advanced encoders capable of extracting richer, context-aware representations through subsequence encoding. For instance, Bidirectional Long Short-Term Memory (BiLSTM) modules are now widely employed to capture long-range sequence dependencies and contextual nuances across entire transcripts [136]. Taking a structural approach, advanced frameworks like GraphLncLoc address the rigid nature of fixed-k features by representing sequences as graphs. This graph-based approach automatically captures patterns and motifs of varying lengths within lncRNA sequences by connecting multiple nodes to form paths. Furthermore, utilizing graph aggregation operations allows these models to cluster neighboring nodes into communities and subgraphs, effectively capturing both global and high-level structural features of the whole lncRNA sequence [144].

## 3. Image-based and Hybrid Methods

In this section, we introduce the state-of-the-art RNA subcellular localization methods using image-based data or both sequence-based and image-based data. Figure 3 summarizes the features and algorithms with different levels of complexity, including machine learning methods, deep learning methods, and complex models that involve multimodal data and multimodal or ensemble learning.

Specifically, from simple to complex approaches, Machine learning methods leverage hand-crafted features extracted from images, such as SIFT, LBP, and HOG, to train models like SVM, RF, and neural networks (e.g., CNN, DNN, GNN, and LSTM). Deep learning approaches further utilize hand-crafted and intensity-based features, including intensity relationships and patterns, to improve predictive performance. Complex models build on these foundations by integrating multi-modal data, such as sequence and image inputs, to create more comprehensive and interpretable models for RNA subcellular localization.

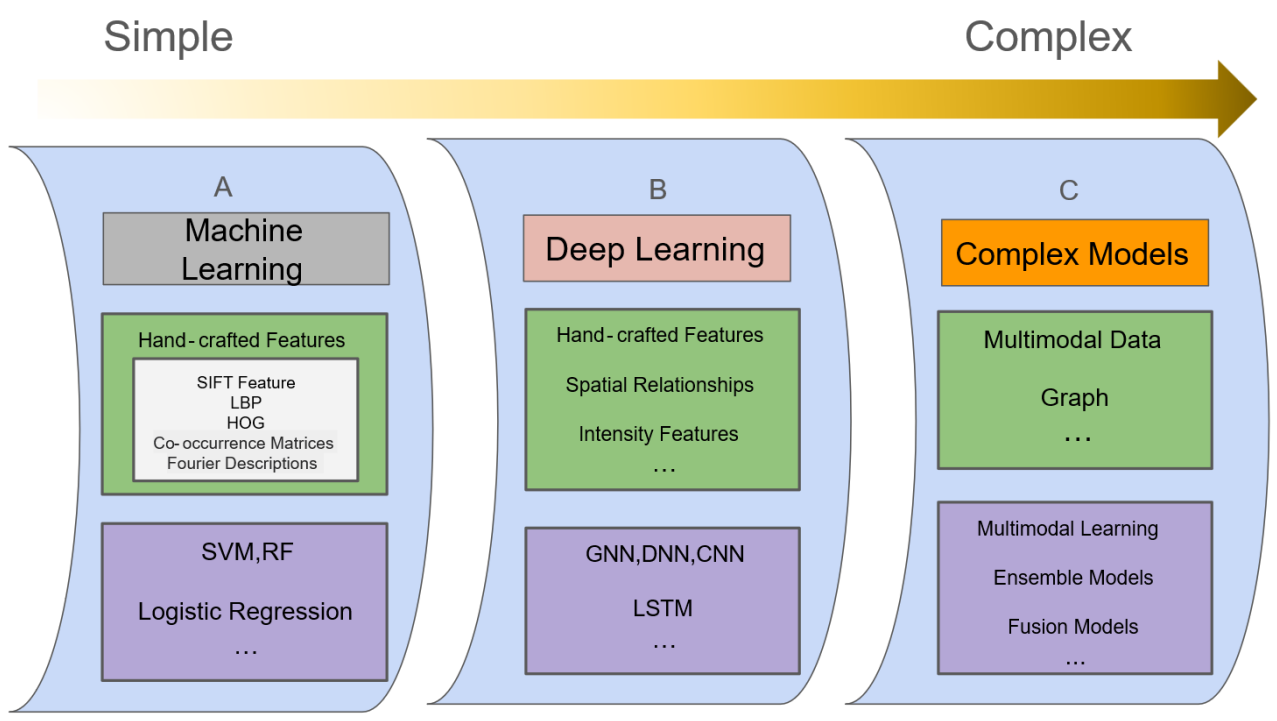

**Figure 3 Three primary categories of computational methodologies for processing imaging data.**
The dark yellow arrow illustrates the increasing complexity of prediction models, representing the progression toward more sophisticated computational frameworks. Green rectangles: features used for model training; purple rectangles: algorithms for location prediction. **(A)** Machine Learning Methods. Hand-crafted features are extracted from images and used to train simple models. **(B)** Deep Learning Methods. This approach employs hand-crafted features, intensity relationships, intensity features, etc. **(C)** Complex Models. This approach combines multi-modality data, such as sequence and image data, as inputs to create a more comprehensive and interpretable model for RNA subcellular localization. SVM (Support Vector Machine), RF (Random Forest), CNN (Convolutional Neural Network), DNN (Deep Neural Network), GNN (Graph Neural Network) LSTM (Long Short-Term Memory), SIFT (Scale-Invariant Feature Transform), LBP (Local Binary Pattern), and HOG (Histogram of Oriented Gradients)

**Table 3** A summary of image-based and hybrid methods for RNA subcellular localization.

| **Method** | **Feature** | **Algorithm** | **Localizations** | **Single / Multiple Labels** | **Reference** | **Year** |
|---|---|---|---|---|---|---|
| Wang et al. | Hybrid | Multimodal fusion models | N/A | N/A | [178] | 2023 |
| Bento | Image (spatial features) | RF | Cell edge; cytoplasmic; nuclear; nuclear edge; random | Multiple | [179] | 2022 |
| Savulescu et al. | Hybrid | Layered neural networks | N/A | N/A | [180] | 2021 |
| Dubois et al. | Image (simulated smFISH pre-processed and down-sampled to 128 x 128 pixels) | CNN (SqueezeNet) | Cell edge; cell extension; foci; intranuclear; nuclear edge; polarized; random | Single | [181] | 2019 |
| Samacoits et al. | Image (curated localization features) | Clustering, RF | Foci; extension; nuclear envelope 2D; nuclear envelope 3D; random | Single | [182] | 2018 |

*Note*: RF: Random Forest; smFISH: single molecule fluorescence in situ hybridization; CNN: Convolutional Neural Network. SqueezeNet is a kind of CNN. In Samacoits et al. [182], nuclear envelope 3D labels are used to discriminate intranuclear distribution and the nucleus membrane that are undisguisable from the view of nuclear envelope 2D. At the time of writing, the hybrid methods [178,180] have not been implemented and so there is no information ("N/A") about localizations or single/multiple labels.

Table 3 summarizes the main image-based and hybrid methods for RNA subcellular localization prediction, ordered chronologically according to their publication years. In the following sections, we give more details about the features and algorithms. To the best of our knowledge, hybrid algorithms like multimodal fusion models [188] and layered neural networks [194] were only proposed but not implemented. Therefore, there is no information about localizations or single/multiple labels, which are indicated by "N/A" in the table.

### *3.1 Image-based Features*

Image-based features are integral to understanding RNA subcellular localization. Unlike sequence-based data, image data from bioimaging and simulation technologies provide spatial and morphological insights, which are essential for determining the precise localization of RNA molecules within different cellular compartments. The image-based features are derived from high-resolution cellular images, enabling researchers to capture and quantify complex biological structures and processes that are invisible to traditional sequencing approaches [183,184]. The application of image-based features in machine learning approaches to RNA localization is facilitated by advancements in microscopy and image processing technologies. In practice, since acquiring ground-truth image data is very expensive and time-consuming, simulated image data for RNA localization patterns were widely used [185,186].

Representative image-based features are key to identifying and understanding the spatial distribution of RNA within cells, which are inspired by the success with image-based mapping of subcellular protein distribution [187]. Raw image data such as smFISH are usually converted into 2D coordinate systems like transcriptomic data formats. Such conversions are helpful for the extraction of spatial features. For instance, the Bento toolkit [179] extracted a total of 13 spatial features that capture the sample's point distribution, including proximity to cellular compartments and extensions, measures of symmetry about a center of mass, and measures of dispersion and point density. More sophisticated spatial-statistical features are also used, such as Ripley's L-functions [188], morphological extraction with RNA counts enrichment ratio [189], and the correlation of z-positions of RNAs and the background intensity [182]. As a more recent trend, raw imaging data are lightly preprocessed before feeding deep neural networks directly, which can automatically extract and learn the most predictive features that capture subtle patterns and variations in RNA distribution that are useful for subcellular localization. Dubois et al.'s work [181] is an example of this trend.

### *3.2 Image-based Algorithms*

To develop machine learning algorithms to predict RNA subcellular localizations from image-based features, the main task is to process and interpret the vast amounts of data they represent. Abundant in the image data are spatial transcriptomics, which were not fully utilized by traditional methods such as RNA sequencing (RNA-seq) or reverse transcription quantitative PCR (RT-qPCR) [178]. As spatial transcriptomics are orthogonal to nucleotide sequence-based data, image-base algorithms are complementary to sequence-based methods in the field of RNA subcellular localization. They not only streamline the analysis of complex image data but also open new avenues for discovering the roles and regulations of RNA and the underlying cellular mechanisms.

Typical image-based algorithms include supervised, self-supervised, and unsupervised techniques. As a robust ensemble supervised learning approach, RF is often chosen to predict RNA localizations from carefully engineered image-based features. In [14], Clarence et al. applied RF to mRNA localization as a multi-label classification problem. Each multi-label classifier consisted of 5 binary classifiers with the same base model, targeting at cell edge, cytoplasmic, nuclear, nuclear edge, and random localizations. In their experiment RF outperformed the popular SVM supervised algorithm and deep learning algorithms such as CNN. However, as a successful deep learning algorithm in computer image recognition [190] and a self-supervised approach, CNN is a preferred algorithm to handle bio-image data in general by automating feature engineering. In [181], for instance, Dubois et al. applied SqueezeNet as a kind of CNN to the classification of 7 mRNA localizations and achieved an overall accuracy of 91%. Unsupervised learning algorithms such as clustering are useful for exploring unlabeled image data and identify clusters as potential RNA subcellular localizations. For example, Samacoits et al. [182] showed how this can be achieved through various clustering methods, including k-means, spectral, and hierarchical clustering.

### *3.3 Hybrid Methods*

As discussed above, conventional machine learning methods are uni-modal, meaning they leverage data of only one modality - either RNA sequence data or image data. While sequence data is a valuable source of sequence patterns, physicochemical features, etc. and image data is a valuable source of transcriptomic and spatial distribution patterns, neither of them alone fully utilize all the patterns encoding RNA subcellular localization. Evaluated by AUC, state-of-the-art algorithms typically fall within the range of 0.4 to 0.8 [131], offering room for improvement.

In recent years, multimodal or hybrid methods are on the rise. In contrast to sequence-only or image-only methods, hybrid methods leverage the complementary strengths of different data sources, facilitating a more holistic understanding of the biological systems involved. For instance, combining sequence-derived features with microscopy images can reveal insights into how sequence motifs are linked to spatial distribution on the subcellular level and improve the accuracy of RNA localization prediction.

The idea of multimodal learning is inspired by the popularity of multimodal deep learning [191], which usually integrates two or more modalities of text, audio, image, and video data, and the success of commercial large language models such as GPT-4o [192]. To utilize the complementary information from multiple modalities, fusion-based deep learning approaches [178] are promising. They have been successfully employed in biomedical fields such as disease prognosis and diagnosis [193], with one such example being breast cancer patient stratification by overall survival [194] when heterogeneous clinical data is available. At the time of writing, the authors were not aware of published multimodal learning algorithms used to predict RNA subcellular localization, although detailed algorithm frameworks have been proposed [178,180]. We envision breakthroughs in applications of hybrid models for RNA localization in near future.

### *3.4 Image-based and Hybrid Methods Progression*

In contrast to the extensive body of literature dedicated to sequence-based models, research utilizing image or hybrid modalities for predicting RNA subcellular localization is scarcer. The development of image-driven and multimodal approaches in this field began significantly later and remains largely immature, leaving many hybrid frameworks at a conceptual or prototyping stage. Nevertheless, these emergent strategies have successfully addressed unique biological dimensions that pure sequence approaches miss. Most notably, image-based methods capture the critical spatial and morphological dynamics of RNA localization within the cellular architecture [182].

Early efforts in this domain primarily utilized traditional Fluorescence In Situ Hybridization (FISH) images. To overcome the limitations of manual curation, some studies attempted to implement simulated frameworks to automate annotations. This shift was motivated by the fact that the manual annotation of 3D point clouds – regardless of their density or reference volume – is exceptionally time-consuming, structurally difficult, error-prone, and highly subjective, particularly when distinguishing subtle localization differences. However, these simulation-based frameworks faced a major bottleneck: they could only annotate previously observed patterns within identified exemplar genes, severely limiting their generalizability [182].

This tension highlights a fundamental limitation of relying solely on genomic sequences. While the functional properties encoded within RNA sequences can explain a large portion of the underlying localization mechanisms, they do not always provide straightforward or sufficient information for predicting actual subcellular localization [178]. Because an organism's primary RNA sequence is invariant, it cannot reflect the highly dynamic nature of cellular prototypes across different cell types, tissue environments, and varying stimulus conditions. Conversely, fluorescence microscopy imaging data serves as the absolute gold standard for determining exact RNA spatial distribution; however, acquiring these images is incredibly costly and technically challenging, making it nearly impossible to scale to the entire transcriptome [181].

To bridge this gap, recent paradigms are shifting toward the integration of multimodal data. Combining diverse modalities, such as sequence-derived features, downstream protein product information, and smFISH-based images, holds substantial potential to provide a far more comprehensive understanding of the multifaceted factors influencing RNA trafficking [180]. By representing these elements as vectors, models can implement advanced feature fusion strategies to synthesize these heterogeneous data streams. While early hybrid attempts relied on simple feature-level fusion, such as concatenating or averaging vectors from different modalities, modern frameworks are increasingly adopting sophisticated token-level and cross-attention fusion mechanisms [178]. These advanced fusion layers allow the model to dynamically weight the importance of sequence motifs against spatial imaging signals, extracting higher-level abstract features that capture the intricate, non-linear interactions among these factors. Ultimately, this holistic representation yields significantly improved and context-aware localization predictions across diverse subcellular compartments.

## 4. Challenges and Future Directions

Despite its remarkable progress, the application of machine learning approaches to RNA subcellular localization also presents many challenges.

One such challenge is the scarcity and quality of data. RNA localization data is often limited, and obtaining datasets that are high-quality, experimentally validated, and comprehensive can be resource intensive. For example, as Wang et al. [188] points out, the genome profile data may lack paired histopathology image data for multimodal deep learning approaches to localization prediction. This scarcity can lead to models that are trained on insufficient or biased data, impacting generalizability and predictive power.

Another significant challenge is the over-reliance on sequence data coupled with the underutilization of image data. Most current ML models heavily depend on sequence data, such as k-mer composition and RNA-protein binding motifs, which, while informative, may not capture the full complexity of RNA localization. The study by Gudenas and Wang [158] highlighted that k-mer composition accounted for 90% of their model's decision-making process, indicating a potential over-reliance on sequence-based features. This narrow focus can overlook other critical determinants of RNA localization such as secondary and tertiary RNA structures, post-transcriptional modifications, and dynamic interactions within the cellular environment.

Although sequence data-based approaches far outnumber image data-based approaches, image data can provide rich spatial and contextual information about RNA molecules within cells, as was discussed in Section 3. Despite advancements in bioimaging techniques, accurate and automated annotation remains a hurdle, which can also be very time-consuming. When there is a lack of high-quality real image data, simulated data such as smFISH are often used [187]. Imperfect segmentation and compartment identification can skew RNA quantification, impacting the reliability of ML predictions.

A further challenge is the lack of large-scale benchmark datasets for all RNA types to ensure replicability of experimental results. Popular open-source datasets such as the RNALocate-based mRNA dataset [114] are often limited in size (<10k samples), restricted to one type of RNA (e.g., mRNA), and biased towards sequence data, in particular nucleotide sequences. The absence of comprehensive, standardized datasets makes it difficult to validate and compare the performance of different ML models. Benchmark datasets are crucial for assessing the robustness and generalizability of ML approaches across various RNA types and experimental conditions.

Another promising future direction is the application of pre-trained RNA language models and foundation models to RNA subcellular localization prediction. By learning contextual representations from large-scale unlabeled RNA sequences, these models can capture complex sequence dependencies and reduce dependency on manually designed features. Recent approaches such as RNALoc-LM [198] have demonstrated the potential of pretrained RNA representations for predicting the localization of multiple RNA types. However, current models remain largely sequence-based and may not capture other important determinants of localization, such as spatial imaging information. As mentioned before, future work could explore multimodal foundation models that integrate sequence representations with structural and imaging data.

On the flip side, these challenges present opportunities for advancement of future directions in this field. Addressing data scarcity through the development of high-throughput experimental techniques and collaborative data-sharing initiatives can significantly enhance model training. Additionally, diversifying the types of features used in ML models – by incorporating structural, biochemical, and interaction data alongside sequence data – can lead to more comprehensive and accurate predictions. As we discussed in Section 3, hybrid methods such as multimodal learning algorithms can utilize the complementary information encoded in both sequences and images, which can offer a holistic view of RNA localization.

While the field of RNA subcellular localization faces significant challenges related to data scarcity, reliance on sequence data, the underutilization of image data, and the lack of large-scale benchmark datasets, these obstacles also present exciting opportunities. Efforts to build standardized datasets or comprehensive databases of RNA localization, such as lncLocator and lncATLAS for LncRNA localization [195], provide valuable resources for the research community. Integrating diverse data types, such as sequence data and image data, can enhance predictive models. Hybrid models such as multimodal learning approaches [188] that combine these data sources have shown promise in improving prediction accuracy and providing deeper insights into RNA localization mechanisms. By embracing a multifaceted approach that combines innovative experimental techniques, comprehensive data integration, and hybrid ML methodologies, researchers can overcome these challenges and unlock the full potential of ML in understanding the complexities of RNA subcellular localization. Table 4 summarizes the major challenges and potential solutions for RNA subcellular localization prediction discussed in this section.

**Table 4** Major challenges and potential solutions for RNA subcellular localization prediction.

| **Challenge** | **Solution** |
|---|---|
| Scarcity of high-quality RNA localization data | Develop high-throughput experimental techniques and promote collaborative data sharing to expand high-quality training datasets |
| Over-reliance on sequence-based features | Incorporate additional structural, biochemical, and RNA-interaction data to provide more comprehensive representations |
| Underutilization and difficult annotation of image data | Improve automated image annotation and integrate spatial imaging information with sequence-based features through multimodal learning |
| Lack of large-scale, standardized benchmark datasets | Establish comprehensive benchmark datasets and databases covering diverse RNA types and subcellular localizations |
| Limited ability of current models to integrate heterogeneous information | Develop hybrid and multimodal models that combine sequence, imaging, and structural data |

## 5. Conclusion

This comprehensive review on RNA subcellular localization has explored three main categories of machine learning approaches: sequence-based, image-based, and hybrid methods. A significant contribution of this work lies in its broad coverage that encompasses various RNA types, including mRNA, lncRNA, miRNA, and other small RNAs. Sequence-based methods utilize RNA sequences to predict subcellular localization, benefiting from the ease of sequence acquisition. Image-based methods leverage bioimaging techniques to analyze RNA localization within cellular compartments, providing a more direct observation of RNA distribution. The integration of sequence and image data in hybrid methods represents a more promising direction, combining the strengths of both data types while eliminating the weakness of either one alone to enhance predictive accuracy by capturing both the intricate details and the bigger picture of RNA sequences and their spatial context within the cell.

The field of RNA subcellular localization prediction still has several key challenges to address, however. The development of more robust and generalizable models requires larger and more diverse datasets, improved feature extraction techniques, and more sophisticated algorithms capable of handling the complexity of biological data. Additionally, there is a need for standardized benchmarks and evaluation metrics to facilitate proper comparison of different methods and ensure reproducibility.

The convergence of machine learning and RNA biology holds great promise for advancing our understanding of RNA subcellular localization. By continuing to refine and integrate computational methods with experimental approaches, we can unlock new insights into the spatial dynamics of RNA, paving the way for novel therapeutic strategies and a deeper understanding of cellular biology.

## Key points

- RNA subcellular localization is central to understanding processes like gene expression regulation with spatial and temporal precision, and AI/ML approaches offer scalable alternatives to labor-intensive manual methods.
- This review makes a comprehensive comparison of sequence-based, image-based, and hybrid methods across mRNA, lncRNA, miRNA, and other small RNAs.
- We critically discuss the challenges in AI/ML approaches for RNA subcellular localization, such as data scarcity and lack of benchmarks, and opportunities to address them.

## Acknowledgements

Research reported in this publication was supported by the U.S. National Science Foundation under Award Numbers 2500836 and 2614824, the National Cancer Institute of the National Institutes of Health under Award Number R03CA317707, the National Institute Of Alcohol Abuse And Alcoholism of the National Institutes of Health under Award Number R21AA032098, and the Office Of The Director, National Institutes Of Health of the National Institutes of Health under Award Number R03OD038391. This research was supported by the State of Nebraska through the Pediatric Cancer Research Group, part of the Child Health Research Institute. This work was also partially supported by the University of Nebraska Collaboration Initiative Grant from the Nebraska Research Initiative (NRI). The content is solely the responsibility of the authors and does not necessarily represent the official views of the funding organizations.

## Author contributions

**Cece Zhang:** Data curation (equal); formal analysis (equal); investigation (equal); methodology (equal); visualization (equal); writing – original draft (equal). **Xuehuan Zhu:** Data curation (equal); formal analysis (equal); investigation (equal); methodology (equal); visualization (equal); writing – original draft (equal). **Nick Peterson:** Writing – review & editing (equal). **Jieqiong Wang:** Conceptualization (equal); project administration (equal); writing – review and editing (equal). **Shibiao Wan:** Conceptualization (equal); funding acquisition (lead); project administration (equal); supervision (lead); writing – review and editing (equal).

## Conflicts of interest

The authors have declared that no competing interests exist.

## Data availability statement

No new data were generated in this study. Data sharing is not applicable to this article.